\documentclass[anonymous]{article}
\PassOptionsToPackage{numbers, compress}{natbib}

\usepackage[table,xcdraw]{xcolor}
\usepackage[preprint]{neurips_2022}

\usepackage[utf8]{inputenc} 
\usepackage[T1]{fontenc}    

\usepackage{url}            
\usepackage{booktabs}       
\usepackage{amsfonts}       
\usepackage{nicefrac}       
\usepackage{microtype}      
\usepackage{xcolor}         

\usepackage{listings}       
\usepackage{graphicx} 
\usepackage{float} 
\usepackage{subfigure} 
\usepackage{amsmath}
\usepackage{mathrsfs}
\usepackage{hyperref}       
\usepackage{algorithm}
\usepackage{algorithmic}
\usepackage{authblk}

\newcommand{\argmax}{\mathop{\arg\max}}
\newcommand{\argmin}{\mathop{\arg\min}}
\newcommand{\ZSEM}{\mathcal{Z}_{sem}}
\title{Latent Magic: An Investigation into Adversarial Examples Crafted in the Semantic Latent Space}

\author{
  BoYang Zheng
}
\affil{\textbf{Shanghai Jiao Tong University}}

\begin{document}

\maketitle

\begin{abstract}
  Adversarial attacks against Deep Neural Networks(DNN) have been a crutial topic ever since \cite{goodfellow} purposed the vulnerability of DNNs. However, most prior works craft adversarial examples in the pixel space, following the $l_p$ norm constraint. In this paper, we give intuitional explain about why crafting adversarial examples in the latent space is equally efficient and important. We purpose a framework for crafting adversarial examples in semantic latent space based on an pre-trained Variational Auto Encoder from state-of-art Stable Diffusion Model\cite{SDM}. We also show that adversarial examples crafted in the latent space can also achieve a high level of fool rate. However, examples crafted from latent space are often hard to evaluated, as they doesn't follow a certain $l_p$ norm constraint, which is a big challenge for existing researches. To efficiently and accurately evaluate the adversarial examples crafted in the latent space, we purpose \textbf{a novel evaluation matric} based on SSIM\cite{SSIM} loss and fool rate.Additionally, we explain why FID\cite{FID} is not suitable for measuring such adversarial examples. To the best of our knowledge, it's the first evaluation metrics that is specifically designed to evaluate the quality of a adversarial attack. We also investigate the transferability of adversarial examples crafted in the latent space and show that they have superiority over adversarial examples crafted in the pixel space. 
\end{abstract}

\section{Introduction}
  Deep Neural Networks(DNN) have achieve high(often state-of-art) performance in various tasks. However, research about the robustness of DNNs \cite{goodfellow} have shown that they are not reliable for out-of-distribution inputs. In image classification tasks, we can reduce the classification success rate to a amazingly low level(sometimes even lower than random) by adding carefully crated noise to the input images. The noise added is confined to a certain degree, often quasi-perceptible by human. The existence of adversarial examples is a big threat to the reliability of DNNs. Therefore, it's crutial to expose as many blind spots of DNNs as possible.

  Many methods have been purposed. The most common method is to add small perturbations to the original images based on direct gradient ascent on the loss function(aka iterative/single step methods). Many iterative/single-step optimization methods have been purposed and achieve high success rate. For examples, BIM \cite{kurakin2016adversarial}, I-FGSM \cite{kurakin2017adversarial}, MI-FGSM \cite{dong2018boosting}, PGD \cite{PGD} have all reached high performance. There are also generator-based methods that train a noise generator against a target model. Results (\cite{gen1},\cite{gen2},\cite{gen3} )have shown that generator-based methods have strong cross-model transferability. However, most prior works manipulate images in the pixel space, which often create spatially regular perturbation to the images. Though the perturbation is confined to a certain degree and is claimed imperceptible, human can still distinguish the pattern of noise in the adversarial examples. A preturbed image is given below as {}, with a common constraint $l_\infty<=16$.

\begin{figure}[htbp]
    \centering
        \centering
        \includegraphics[width=0.8\linewidth]{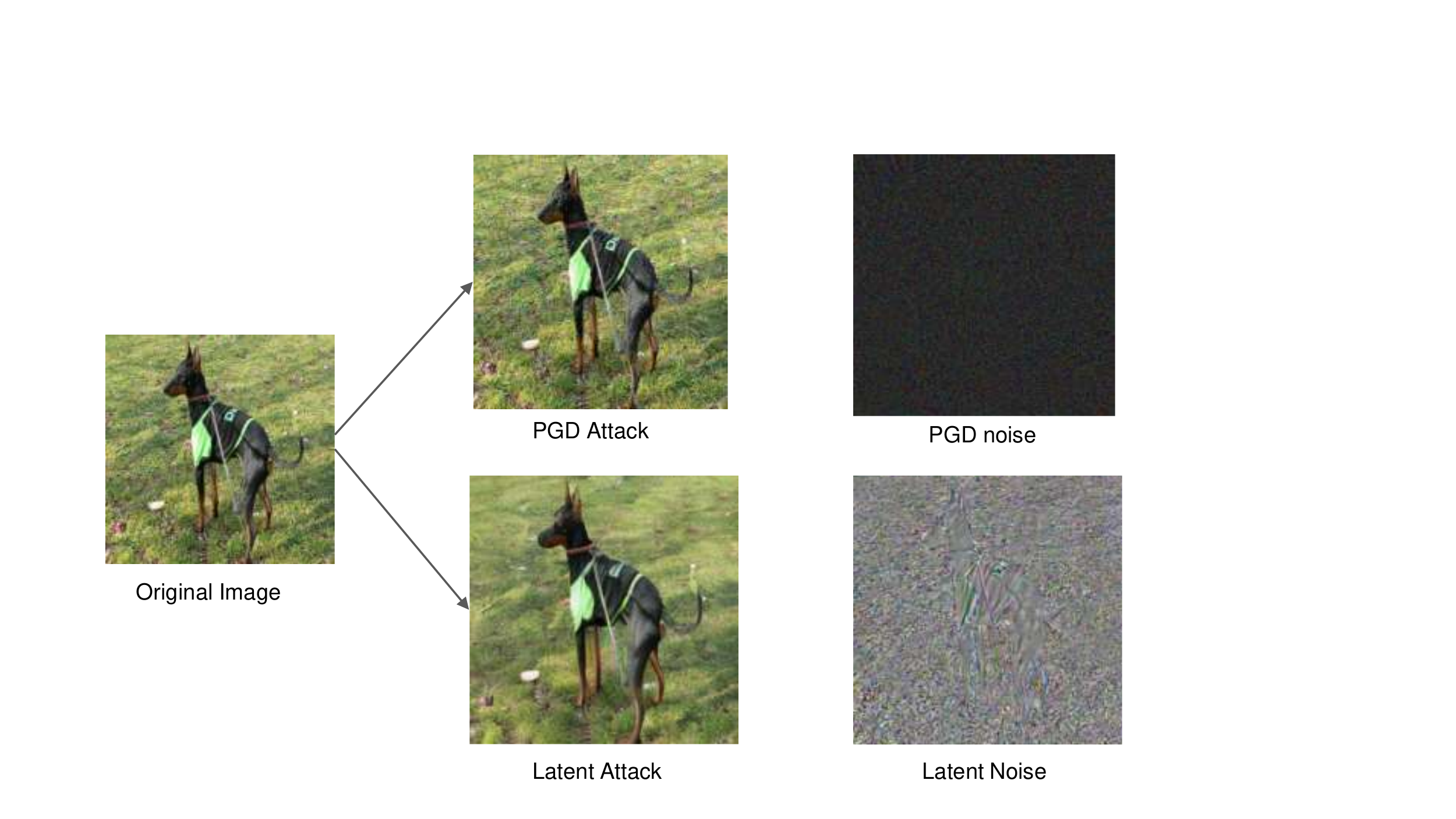}
    \centering
    \caption{Comparison between adversarial examples crafted in pixel space and latent space. Under latent attack, the perturbation is more covert, and the noise is highly semantic. PGD attack is under a noise budget of $l_\infty<16$}
    \label{fig:compare}
  \end{figure}

  Rather than noising images directly in the pixel space, some works have explore adding noise beyond traditional methods. AdvGAN++\cite{ADVGAN++},ATGAN\cite{ATGAN} explore the possibility to add noise in the down-sampled space.\cite{NATUREADV} explicitly search adversarial examples in the latent space created by GAN. \cite{DataPoisonGen} purpose to find adversarial examples in the latent space created by Variational Auto Encoder(VAE). However, we argue that these methods are purposed before a large, pre-trained encoder-decoder model exists. Thus the latent space they generated are relatively coarse and the adversarial examples they crafted are not as effective as those crafted in the pixel space. A more recent work \cite{SDM} train a VAE model to project images into a latent space as part of the Stable Diffusion Model. Results in \cite{SDM} shows the latent space created by the model is highly semantical and precise. We take advantage of them by utilizing their VAE as our pre-trained encoder-decoder structure. Therefore, we're able to make use of the semantic latent space and create purturbed images. In section \ref{results} , we show that our method can achieve a comparable attack rate as those crafted in the pixel space, while our noise being much more imperceptible to human. A comparison is given below as figure \ref{fig:compare}.

\begin{figure}[htbp]
      \centering
      \includegraphics[width=\linewidth]{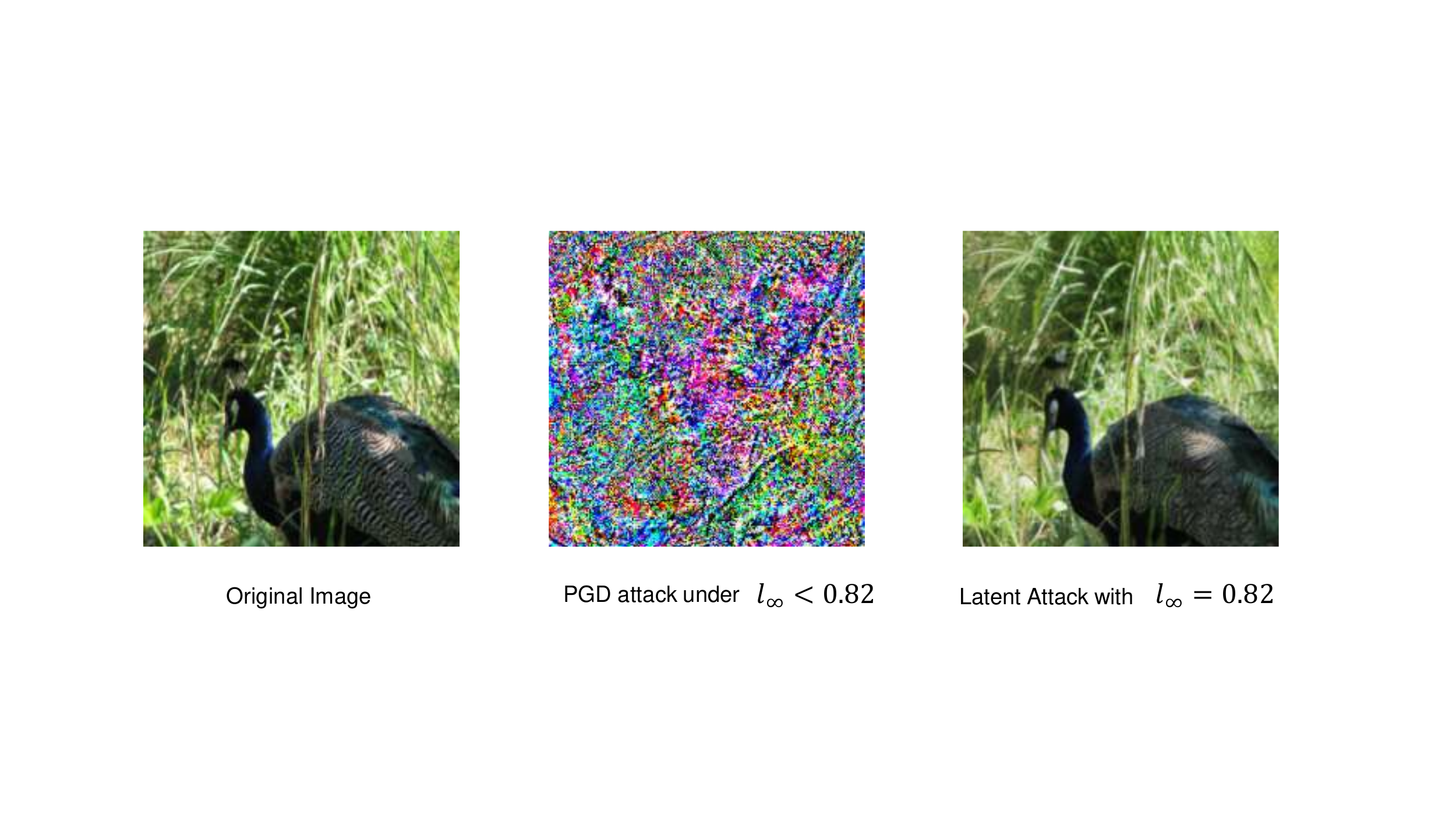}
      \caption{A adversarial example crafted from latent space. The perturbation is almost imperceptible, comparing to a normal PGD attack with $l_\infty=0.82$ }
      \label{0.82}
  \end{figure}

  However, as shown in \cite{Concurrent}, preturbed images decoded from latent space are extremely hard to follow the common $l_p$ norm constraint. As illustrated in figure \ref{0.82}, the perturbation is almost imperceptible while the $l_\infty$ distance is larger than the normal constraint. Thus a new evaluation metric is needed for the adversarial examples crated from latent space. Existing methods includes FID score \cite{FID}, Structural Similarity Index Measure (SSIM) and Peak Signal-to-Noise Ratio(PGNR)\cite{genlatenttrash}. In section\ref{analyze_metric}, we discuss the weakness of those pre-existing evaluation metrics, and purpose our novel metric based on SSIM loss and the fool rate. To the best of our knowledge, our work is \textbf{the first} to systemetically explore the quality of adversarial attack on latent space.

  As we are disturbing images in a semantic level, we expect the perturbation to have a high level of transferability. Thus in section\ref{transfer} we investigate the cross-model transferability of adversarial examples crafted from semantic latent space. The results show that adversarial examples crafted from latent space have a high level of transferability, outperforming the transferability of adversarial examples crafted by \textbf{PGD} and FGSM. 

  We summarize our contributions as follows:

  \begin{itemize}
    \item We give explanations about the reason to noise the images in the semantic latent space. Base on a pre-trained VAE from Stable Diffusion Model, we show that adversarial examples crafted from latent space can achieve a comparable fool rate as those crafted in the pixel space, while the perturbation being more imperceptible.
    \item  We purpose a novel evaluation metric for adversarial examples crafted from latent space. \textbf{To the best of our knowledge, our method is the first metric specifically designed for adversarial examples crafted from latent space.}
    \item We investigate the cross-model transferability of adversarial examples crafted from semantic latent space. We also investigate how the choice of loss funtions affects the transferability.
  \end{itemize}
\section{background}
\subsection{Iterative Attack Methods}
  In order to generate adversarial examples, many methods have been purposed. The most common method is to add small perturbations to the original images based on direct gradient ascent on the loss function(aka gradient-based methods). In the white box setting, many iterative optimization methods have been purposed and achieve high success rate. For examples, BIM \cite{kurakin2017adversarial}, I-FGSM \cite{kurakin2016adversarial}, MI-FGSM \cite{dong2018boosting}, PGD \cite{PGD} have all reached high performance.
\subsection{Variational Auto Encoder and Stable Diffusion Models}
  Variational Auto Encoder(VAE)\cite{VAE} is an encoder-decoder structure generative models for generating images. The encoder encodes image to a latent variable and the decoder decodes it back into pixel space. The sub-space that encoder encodes image into is called the latent space. As our research manipulate latent variables on latent space, the outcome of our experiments will strongly depend on a well-trained VAE that creates a proper latent space. After some pilot runs, we adopted the pre-trained VAE from Stable Diffusion Model\cite{SDM} as our VAE module. Stable Diffusion Model is Latent Diffusion Model that use a U-Net to recurrently noise and denoise the latent variable image in the latent space. \cite{SDM} purpose that the latent space created by LDMs eliminates imperceptible details of original images while maining the semantic information. The outstanding achievement of stable diffusion models strongly supports the authors' claim. Therefore, we adopt the pre-trained VAE from Stable Diffusion Model to create a strong semantic latent space for our research.
\section{Related Works}
  \subsection{Adversarial Attack in the latent space}
  \cite{DataPoison} first purpose to generate adversarial examples by noising the latent variable created by VAE. They tends to learn a constant noise $\Delta z$ for all the latent variables. Later, an incremental work \cite{DataPoisonGen} trains a generator in the latent space to produce noised latent variables. \cite{AVAE} purpose AVAE, a model based on VAE and GAN\cite{GAN} to produce adversarial examples.  And \cite{NATUREADV} purposed to search for latent variables in a latent space created by GAN and create semantical change to the image. The forementioned works are in lack of quantitive analysis of the quality of the adversarial examples and the explanation of the reason to noise the latent space.
  
  Notably, a concurrent work \cite{Concurrent} use Latent Diffusion Model to produce noised images, which also uses the same latent space as ours. They purpose to perform 5 steps of denoising process of DDIM\cite{DDIM} on the latent variable $z=\mathcal{E}(X)$ before decoding. They tends to maximizing the loss by updating $z$, which is much similar to ours method.

\begin{figure}[htbp]
      \centering
      \includegraphics[width=\linewidth]{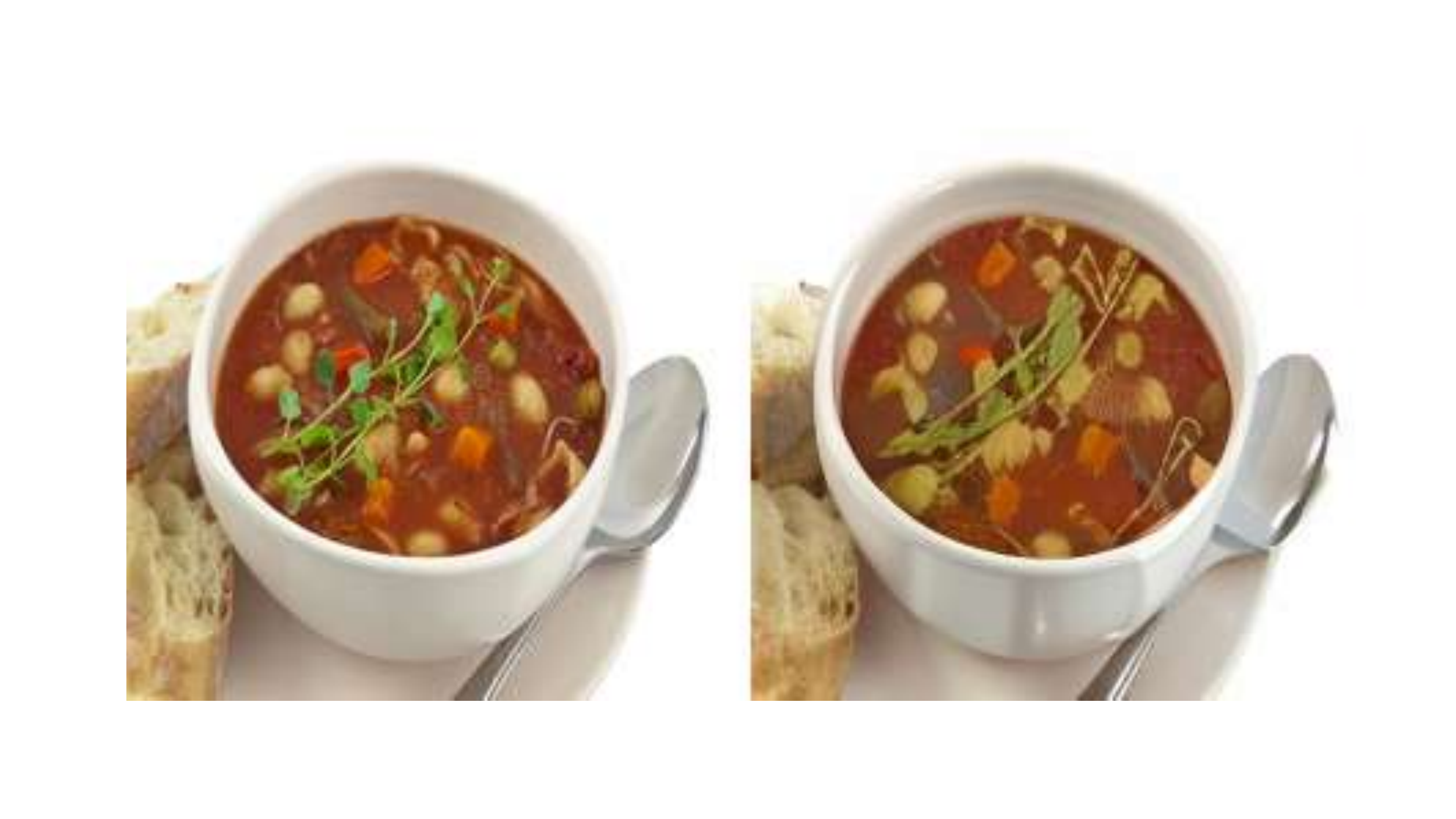}
      \caption{The image on the left is the original images, and the image on the right is produced using the open-source code of \cite{Concurrent} by their default settings. As illustrated, the denoising process of DDIM has totally change the vegetable in the middle, which is a huge semantic change. The denoising process also purify the watermark on the background, which is also not expected.  }
      \label{soup}
  \end{figure}

  However, we claim that though they take the structure into account by adding self-attention and cross-attention loss, the denoising step could still change the semantic information to a human-perceptible extent. An example is given as figure \ref{soup}. As their works manipulate the semantic meaning of the image on a high level that's visible to human, we argue that the denoising process of DDIM is too strong a perturbation to be used upon the latent variable. Thus we won't compare with their work in the experiment section, even though we reach comparable transferability as theirs.

  We admit that our basic framework for crafting adversarial examples in the latent space is not novel and has much similarity with previous work. However, the explanation for the reason to noise the latent space and the evaluation metric we purpose are novel and important contributions of our work. Our experiments also illustrate the high transferability of adversarial examples crafted from latent space, which is a novel finding.
\section{Problem Formulation}
We denote the target classification model as $ \mathcal{F} $ . For the VAE we used, we seperately denote the encoder and the decoder as $\mathcal{E}$ and $\mathcal{D}$. We denote the latent space as $\ZSEM$, our loss function as $\mathcal{L}$

In the noising phase, we assume to know every information about $\mathcal{F}$. For a given dataset $X$, we first encode $X $ into a latent variable $z=\mathcal{E}(X)\in \ZSEM$ 

The noising goal is to maximize:

$$
z' \longleftarrow \argmax_{z'} \mathcal{L}(\mathcal{F}(\mathcal{D}(z')))
$$

Then the noised latent variable $z'$ is decoded back into pixel space as $X'=\mathcal{D}(z')$. The adversarial example is then generated.

\section{Why Latent Space?}
Though crafting adversarial examples is a crutial topic in the field of adversarial machine learning, researches about crafting in the latent space is still relatively less explored. In this section, we will give explanations about the reasonabilityand neccessity  of crafting adversarial examples in the latent space.
\subsection{Comparison with Generator-Based Methods in the Pixel Space}
Many researches have been done on training a noise-generator to generate adversarial examples. The most common adversarial generator model structure consists of a down-sampler, a bottle-neck module and an up-sampler. We denote the down-sampler as $\mathcal{D}$, the up-sampler as $\mathcal{U}$, the bottle-neck module as $\mathcal{B}$, then the generator can be denoted as $\mathcal{G}=\mathcal{U}\circ\mathcal{B}\circ\mathcal{D}$. If we view $\mathcal{D}$ and $\mathcal{U}$ as an encoder-decoder structure model, then $\mathcal{G}$ is a generator that produces noises in the latent space created by $\mathcal{U}$ and $\mathcal{D}$. In fact, ATGAN\cite{ATGAN} has explicitly purposed to add perturbation on the down-sampled space. Thus, we argue that the traditional generator-based methods can be viewed as a special case of our method, where the latent space is created by $\mathcal{U}$ and $\mathcal{D}$. However, as $\mathcal{U}$ and $\mathcal{D}$ is often trained under the limitation of $l_p$ norm, the latent space created by them is less semantic than a $l_p$ norm-free latent space but contains more spatial details. 
\subsection{Making Use of the Feature of Human Recognition System}
The main difference between the latent space and the pixel space is that perturbations made on latent space cause semantic changes while those made on pixel space cause changes with spatial pattern. We argue that under the same extent of change, the Human Recognition System is more tolerant to semantic change rather than the spatial.

\begin{figure}[htbp]
    \centering
    \includegraphics[width=\linewidth]{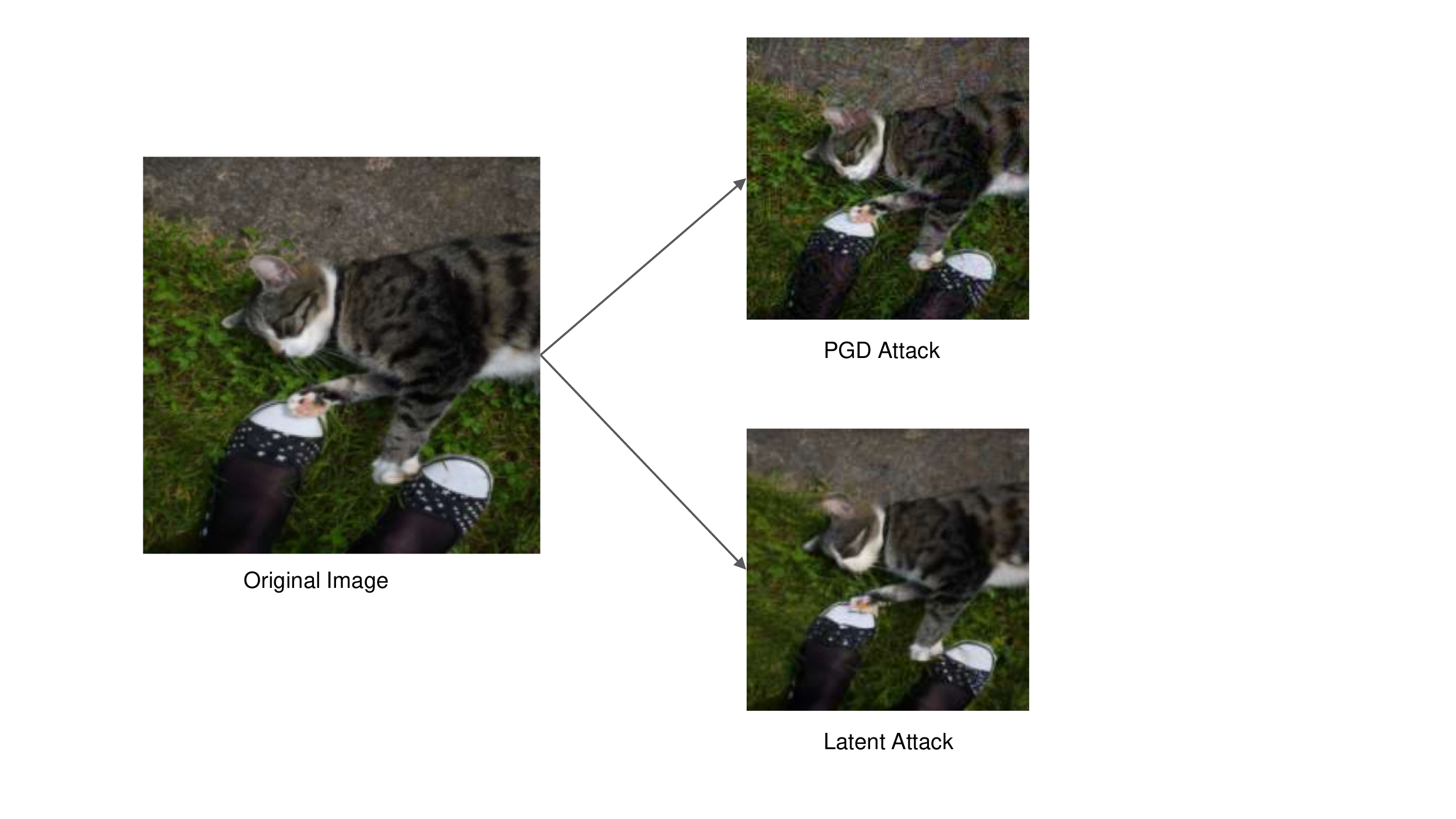}
    \caption{An illustration of the robustness of Human Recognition System to semantically out-of-box distribution. The PGD attack is under the $l_\infty$ norm constraint with $l_\infty<16$. The noise produced by PGD as colorful stripes can be easily seen by human, while the noise produced by our method is semantically more natural. Please zoom in to see the details.}
    \label{catty}
\end{figure}

It's widely acknowledged that human recogizes images in a semantic-level. For example, the original image in figure \ref{catty} would be recogized by human as cat, a shoe and grass ground, along with their interactions with each other. It's shown in \cite{CLIP} that the CLIP model can encoded the semantic information of an image into a latent variable with strong semantic information, which partially proves that a feature extractor is able to extract semantic information from an image as well as human does. However, neural networks have been proved to have linear nature \cite{goodfellow} in the feature space, while human recognition system is strongly non-linear. As illustrated in figure \ref{catty}, a slightly semantically-preturbed cat would still be recogized as cat to human. But actually it is an unnatural being and should never been seen before, as no cat will have a strange pow and be without mouth. \textbf{We believe the core reason behind this scenario is that human kept a strong robustness to semantically out-of-box distribution, while neural networks are easily fooled due to their linear nature.}

However, human are much more sensitive to spatial change. We can easily recogize the spatial pattern of the noise, as to human the noise is a new semantical object that's independent to other objects in the image. Meanwhile, neural networks are equally easy to be fooled by both kind of change. Thus, it's reasonable for attackers to manipulate images by making semantical changes rather than spatial changes, as it's more covert and imperceptible to human beings.

\section{Similarity-Delta Measure($SDM$):\\ A Metric for Evaluating Adversarial Attacks on Latent Space}
During practice, we found a lack of evaluation methods for adversarial attacks on the latent space. We argue that most of the existing evaluation methods are not suitable for evaluating adversarial attacks on the latent space and give our detailed analysis on existing metrics. After, we purpose a novel metric called $SDM$ to quantitively measure the quality of a latent space adversarial attack.
\subsection{Analyzing Existing Evaluation Metrics}\label{analyze_metric}

Unlike the perturbation in the pixel space, the perturbation in the latent space hardly follows the $l_p$ norm constraint. Some other metrics have been used to evaluate the adversarial examples crafted from latent space.
We analyze those metrics and provide reasons for why they are not suitable to evaluate adversarial attacks on latent space.  
\subsubsection{Deep Learning Based Metrics}
The most famous deep learning based metric is Frechet Inception Distance score(FID)\cite{FID}. FID is a metric to evaluate the similarity between two set of images based on the feature extracted from the Inception model\cite{Inception}. The lower the FID score is, the more similar the two set of images are. FID is widely used in GAN\cite{GAN} and VAE\cite{VAE} to evaluate the quality of the generated images. However, we claim that FID is not suitable for evaluating adversarial examples. We take a special situation as example, where the adversarial example is crafted against the Inception model, which is also used to extract the feature in FID scoring. In this situation, the adversarial example is attacking the feature extractor itself, thereby affecting the features. Thus, the FID score will not be the actual perception distance between the adversarial example and the original image. 

From a broader perspective, for other deep learning based metrics(like lpips\cite{LPIPS} ), what they actually evaluate is the distribution similarity\cite{FID} between two sets of images. However, the adversarial example is often designed to be perceptually similar to the original image, but is actually out of the original distribution. Thus from the nature of deep learning based metrics, they are unfair for adversarial examples.

\subsubsection{Traditional Metrics}
For traditional metrics, the most commonly used ones are Structural Similarity Index Measure($SSIM$)\cite{SSIM} and  Peak Signal-to-Noise Ratio($PSNR$)\cite{PSNR}. $SSIM$ is a metric to evaluate the similarity between two images. $PSNR$ is a metric to evaluate the quality of the image. Both of them are widely used in image processing. However, only reporting those losses is not enough, as lower $SSIM$ and higher $PSNR$ could allow more perturbation to the images, which often indicates a better attack rate. Thus a single loss without baseline is meaningless. Meanwhile, an universal baseline for such losses is hard to measure. For example, different datasets could have a different baseline of $SSIM$ and $PSNR$, which makes cross-dataset comparison unfair. Even for the same dataset, different encoder-decoder model could make the baseline different. 

Particularly we give more detailed introduction about $SSIM$, as it will be used as our base metric for evaluating adversarial attacks. $SSIM$ is defined as:

$$
\operatorname{SSIM}(\mathbf{x}, \mathbf{y})=\frac{\left(2 \mu_x \mu_y+C_1\right)\left(2 \sigma_{x y}+C_2\right)}{\left(\mu_x^2+\mu_y^2+C_1\right)\left(\sigma_x^2+\sigma_y^2+C_2\right)}
$$

where $\mu_x$ and $\mu_y$ are the mean of $x$ and $y$ respectively, $\sigma_x^2$ and $\sigma_y^2$ are the variance of $x$ and $y$ respectively, $\sigma_{xy}$ is the covariance of $x$ and $y$, $C_1$ and $C_2$ are two variables to stabilize the division with weak denominator. SSIM is a value between 0 and 1, where 1 means the two images are identical. 

In practice, $SSIM$ is often applied to a sliding window of size $w \times w$ on both image, and takes the mean of all the $SSIM$ values as the final $SSIM$ score. We also add a gaussian filter to the sliding window to make the $SSIM$ score more robust to the noise. For implementation details please check appendix {}.

\subsection{$SDM$:Similarity-Delta Measure}
\subsubsection{Similarity Functions}
Though using SSIM or PSNR as metrics is not adaquate, we still believe that traditional similarity functions are good metrics to evaluate the adversarial examples and can be used for our new metric. We let our chosen similarity function as $\mathcal{S}$.
For $S$ we make the following assumptions:
\begin{itemize}
  \item $\mathcal{S}$ is a function that takes two images as input and outputs a value between 0 and 1, where higher value means the two images are more perceptually similar.
  \item $\mathcal{S}$ is symmetric, which means $\mathcal{S}(X, X') = \mathcal{S}(X', X)$
  \item $\mathcal{S}(X,X')=1$ if and only if $X=X'$ 
\end{itemize}
Many similarity functions satisfy the forementioned properties. Additionally, we assume that stronger adversarial examples lead to lower $\mathcal{S}$. And we want to find a balance between the adversarial examples' strength and perceptional similarity to the original image, while being data-independent and model-independent. Thus we propose a new metric based on SSIM, which is Similarity-Delta Measure(SDM). $SDM$ is defined under a fixed dataset $X\sim D_X$ and an encoder-decoder model $\mathcal{M}$.

Additionally, we denote $Acc$ as the accuracy of the adversarial examples crafted from $X$ and $\mathcal{S}$ as the $\mathcal{S}$ score of the adversarial examples.

$Acc_{\mathcal{M}}$ is the baseline accuracy, defined as :

$$
Acc_{\mathcal{M}}= \mathbb{E}_{X \sim \mathcal{D}_{X}}\left[\mathcal{F}(X, \mathcal{D}(\mathcal{E}(X))=labels)\right]
$$

$\mathcal{S}_{\mathcal{M}}$ is the baseline $\mathcal{S}$ score on model $\mathcal{M}$, defined as:

$$
\mathcal{S}_{\mathcal{M}}= \mathbb{E}_{X \sim \mathcal{D}_{X}}\left[\mathcal{S}(X, \mathcal{D}(\mathcal{E}(X)))\right]
$$

Then we define: 

$$\Delta Acc=\frac{Acc_{\mathcal{M}}-Acc}{Acc_{\mathcal{M}}}$$

which indicates the fooling capability of the adversarial examples. Similarly, we define:

$$\Delta \mathcal{S}=\frac{\mathcal{S}_{\mathcal{M}-\mathcal{S}}}{\mathcal{S}_{\mathcal{M}}}$$

which indicates the decay of similarity between adversarial examples and original images.

\textbf{The $SDM$ is defined as:}

$$
SDM= \frac{\log(1-(1-\gamma)* \Delta Acc)}{\Delta \mathcal{S}+\varepsilon}
$$

where $\varepsilon, \gamma $ are small numbers to prevent undefined math operations.

We set $\gamma = 1e-3$ and $\varepsilon=1e-7 $ in our experiments.

To evaluate traditional adversarial attacks, we set $\mathcal{S}_\mathcal{M}=1$.
\subsection{Intuition behind the Form of SDM}
As our goal is to let $SDM$ be a metrics that balance the adversarial examples' strength and perceptional similarity to the original image, it's natural to make a trade-off between $Acc$ and $\mathcal{S}$ by using a ratio of them. However, simply use $SDM=\frac{Acc}{\mathcal{S}}$ is not model and data independent, as the absolute value of $Acc$ and $\mathcal{S}$ strongly relys on the model and dataset. Thus we use the ratio of the relative value of $Acc$ and $\mathcal{S}$ to make $SDM$ model and data independent. The first version of $SDM$ is defined as:

$$
SDM=\frac{\Delta Acc}{\Delta \mathcal{S}}
$$

\begin{figure}
  \centering
  \label{fig:ssim_acc}
  \begin{minipage}[t]{0.45\linewidth}
  \centering
  \includegraphics[width=0.8\textwidth]{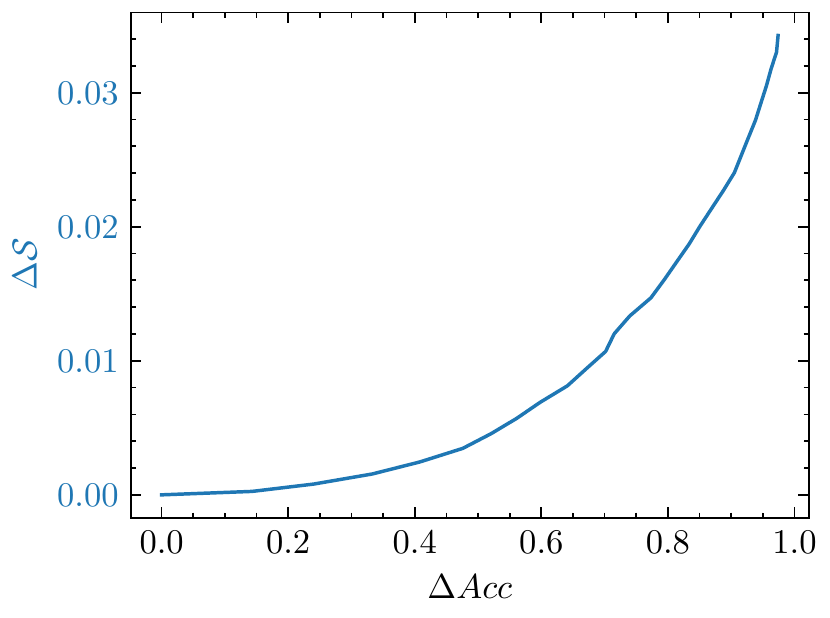}
  \caption{target model is ConvNext\_Base.}
  \end{minipage}
  \begin{minipage}[t]{0.45\linewidth}
  \centering
  \includegraphics[width=0.8\textwidth]{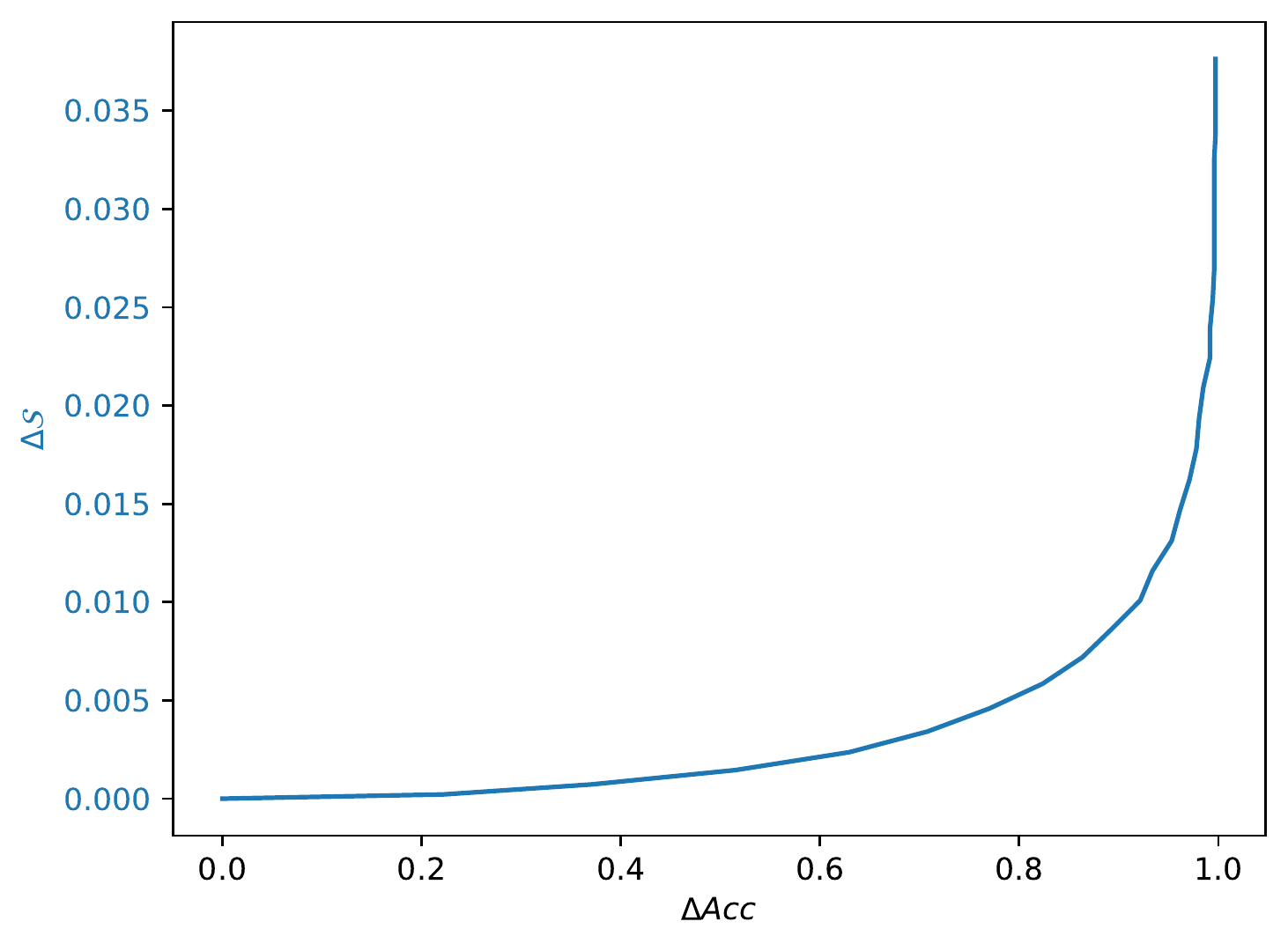}
  \caption{target model is ViT\_Base.}
  \end{minipage}
  \caption{The relation between $\Delta Acc$ and $\Delta S$ }
\end{figure}

However, from practice we observe that $\Delta S$ does not have a linear relation with $\Delta Acc$. As $\Delta Acc$ rises, $\Delta S$ rises more and more rapidly, as illustrated in figure \ref{fig:ssim_acc}. 

It's clear that the relation between $\Delta Acc$ and $\Delta S$ is not linear. Thus we needs to find a proper function $f$ to make the relation between $f(\Delta Acc)$ and $\Delta S$ linear(or constant). Intuitionally, the attack rate is harder to increase when it is closer to 1. Thus we must pay a greater decrease in $\mathcal{S}$ as the "price". We can formulate this by a differential equation:

$$
\frac{d (\Delta Acc)}{d (\Delta S)} = f(1-\Delta Acc)
$$

In this paper, we assume $f$ to be an linear function, where means $f(x)=K x$. Thus we can solve the differential equation and get:

$$
\Delta \mathcal{S}= \frac{1}{K} \log(1-\Delta Acc)
$$

We notice that the value of $K$ is important, as it determines the slope of the curve of $\Delta Acc$ and $\Delta S$, which represents the strength of an adversarial attacks. If $K_1>K_2,\Delta Acc_1=\Delta Acc_2 $, we have $\Delta \mathcal{S}_1< \Delta \mathcal{S}_2 $. Thus a larger $K$ indicates that the adversarial attack is more semantically similar to the original image than other attacks under the same attack rate, which reflects the strength of the adversarial attack. Therefore, we set $SDM=K$. Then we get:

$$
SDM= K= \frac{\log(1-\Delta Acc)}{\Delta \mathcal{S}}
$$

To avoid the denominator to be zero, we add a small number $\varepsilon$ to the denominator. To avoid the numerator to be undefined when $\Delta Acc=1$, we add a small number $\gamma$ to the numerator. Thus we get the final form of $SDM$:

$$
SDM= \frac{\log(1-(1-\gamma)\Delta Acc)}{\Delta \mathcal{S}+\varepsilon}
$$

In practice, we found $SDM$ of this form is more stable to various attack rate, further analysis is given in section \ref{stable_sdm}.
\subsection{Model-Independency and Dataset-Independency of SDM }
We claim that SDM is strongly model-independent and dataset-independent. Though the absolute value of $Acc$ and $\mathcal{S}$ can differ dramatically, by comparing with the baseline of the given model and datasets, we can still get a fair cross-model and cross-dataset comparison.
Below we give detailed analysis about the validity of the baseline chosen.

We claim that a well-trained encoder-decoder structure should satisfy the following properties:

(1) The encoder-decoder is trained to minimize the reconstruction loss $L_{recon}$,which should be equivalent to maximize some similarity function $\mathcal{S}'$. We assume $S'$ also satisfy the forementioned properties of similarity functions. The training goal of the encoder-decoder can be formulated as:

$$
\begin{aligned}
    \argmin_{\mathcal{E}, \mathcal{D}} \mathcal{L}_{\text {recon }} & = \argmax _{\mathcal{E}, \mathcal{D}} \mathbb{E}_{X \sim \mathcal{D}_{X}}\left[\mathcal{S}'(X, \mathcal{D}(\mathcal{E}(X)))\right] \\
    &=\argmax _{\mathcal{E}, \mathcal{D}} \mathbb{E}_{X \sim \mathcal{D}_{X}}\left[\mathcal{S}'(X, X')\right]
\end{aligned}
$$

(2) If the encoder-decoder structure is {\em perfect}(with zero loss on any data), then the $\mathcal{S}$ score of the original image and the reconstructed image should be 1(which means they are identical). Meanwhile, any other image should have a $\mathcal{S}$ score less than 1. That can be formulated as:

$$
    \begin{cases}
        \mathcal{S}(X, X') = 1 & \text{if } X'=\mathcal{D}(\mathcal{E}(X))\\
        \mathcal{S}(X', X') < 1 & \text{otherwise} 
    \end{cases}
$$

(3) However, it's impossible for an encoder-decoder model to be {\em perfect} while maintaining a highly semantical latent space. But we claim that for a well-trained encoder-decoder model, the maximum $\mathcal{S}$ score could only be achieved by the image that's decoded from the encoded image of itself. That can be formulated as:

$$
\begin{aligned}
  \text{Assume }\max\limits_{X' \in \mathcal{D}(\ZSEM)} (\mathcal{S}(X, X')) = C \leq 1 \\
  \text{Then } \mathcal{S}(X, X') = C \text{ if and only if } X'=\mathcal{D}(\mathcal{E}(X))
\end{aligned}
$$

We give a intuition of why this property holds. Assume:

$$\exists z\in \ZSEM,z\neq \mathcal{E}(X),\mathcal{S}(\mathcal{D}(z),X) > \mathcal{S}(\mathcal{D}(\mathcal{E}(X)),X)  $$

Due to the property (1) of similarity functions, $\mathcal{S}'(\mathcal{D}(z),X)>\mathcal{S}'(\mathcal{D}(\mathcal{E}(X)),X)$ should also holds, which contradicts with property (1). Thus such $z$ could not exist. 

However, it's indeed possible to have $z\in \ZSEM$ with $S(\mathcal{D}(z),X)=\mathcal{S}(\mathcal{D}(\mathcal{E}(X)),X)$, but in real world it's nearly impossible to find such $z$ that happens to be exactly identical.

Given the forementioned properties of a well-trained encoder-decoder structure, we can now validate $\mathcal{S}_{\mathcal{{M}}} $, as any perturbation on the original image should lead to a lower $\mathcal{S}$ score. Thus:

$$
\mathcal{\mathcal{M}}-\mathcal{S}+\varepsilon > 0
$$

As for $Acc$, if the accuracy of the preturbed image is higher than the baseline, then the perturbation is not adversarial. Thus for a valid adversarial attack, we have:

$$
Acc_{M}-Acc \geq 0
$$

\textbf{Thus we have some properties of the SDM score}:

\begin{itemize}
  \item $SDM \geq 0$ for a well-trained encoder-decoder model and a valid adversarial attack on latent space.
  \item $SDM$ is larger when adversarial are stronger and more human imperceptible.
  \item $SDM$ is highly model-independent and dataset-independent.
\end{itemize}

\subsection{Choice of Similarity Function:$SSIM$}\label{stable_sdm}
Though $SDM$ is highly model-independent and dataset-independent, it's still sensitive to the choice of similarity function. In figure \ref{fig:ssim_acc} a strong correlation between the accuracy and $SSIM$ is shown. Thus we choose $SSIM$ as the similarity function in our experiments. We also show that use $SSIM$ as $\mathcal{S}$ brings a high level of stability to the $SDM$ metric.

\begin{figure}[htbp]
    \centering
    \includegraphics[width=\linewidth]{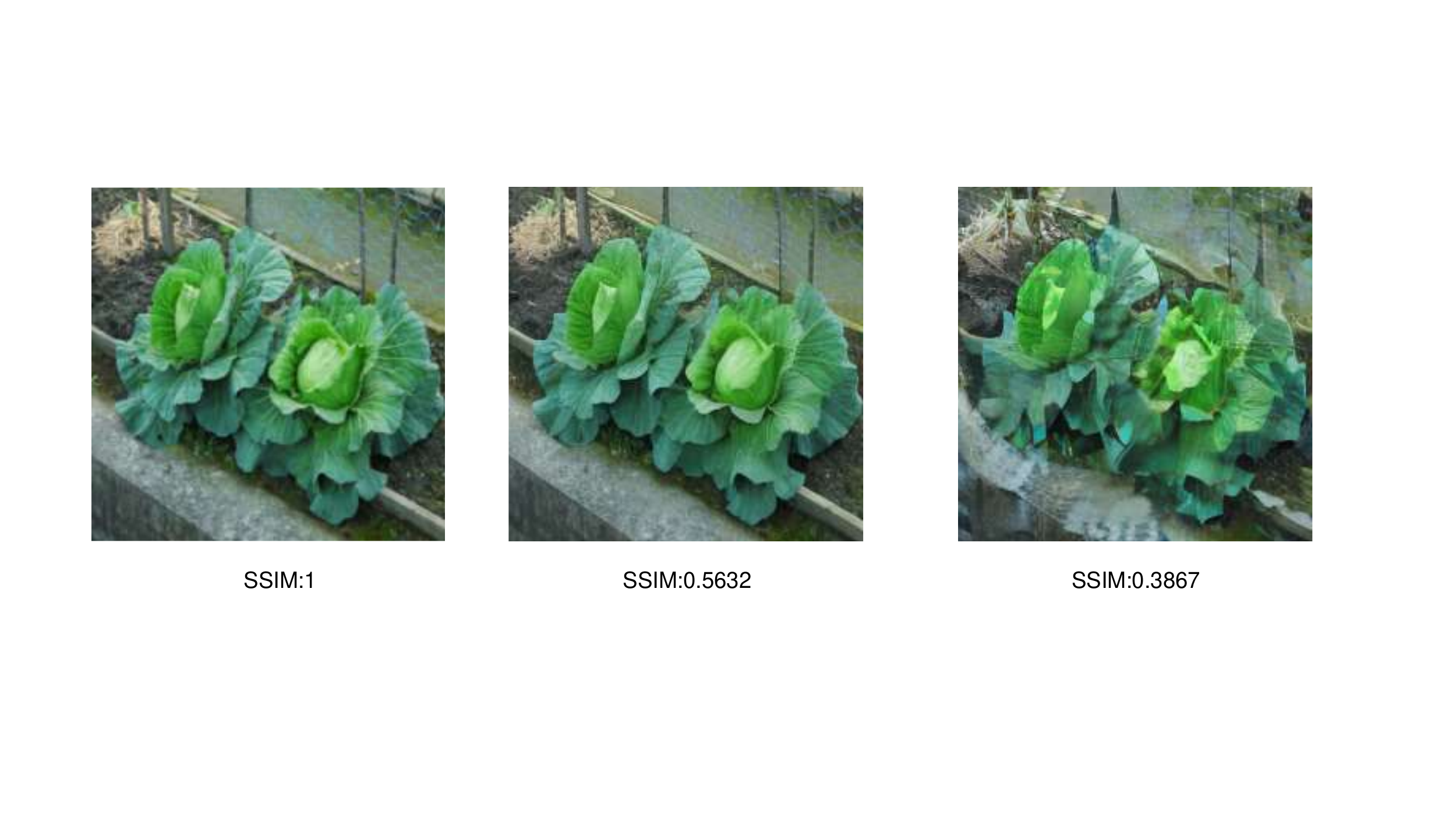}
    \caption{Adversarial examples with different $SSIM$ scores. When $SSIM$ score is low, the perturbation is more human imperceptible.}
    \label{cabbage_ssim}
\end{figure}

To show how $SSIM$ evaluates the image, we craft two adversarial examples from a same image in the latent space, but with different extent of perturbations. The first one have a $SSIM$ score of $0.3627$, the second one have a $SSIM$ score of $0.5652$. As illustrated in figure \ref{cabbage_ssim}, we can clearly see the difference between purturbed images grows as the $SSIM$ score lowers.

\begin{figure}[htbp]
    \centering
    \includegraphics[width=0.6\linewidth]{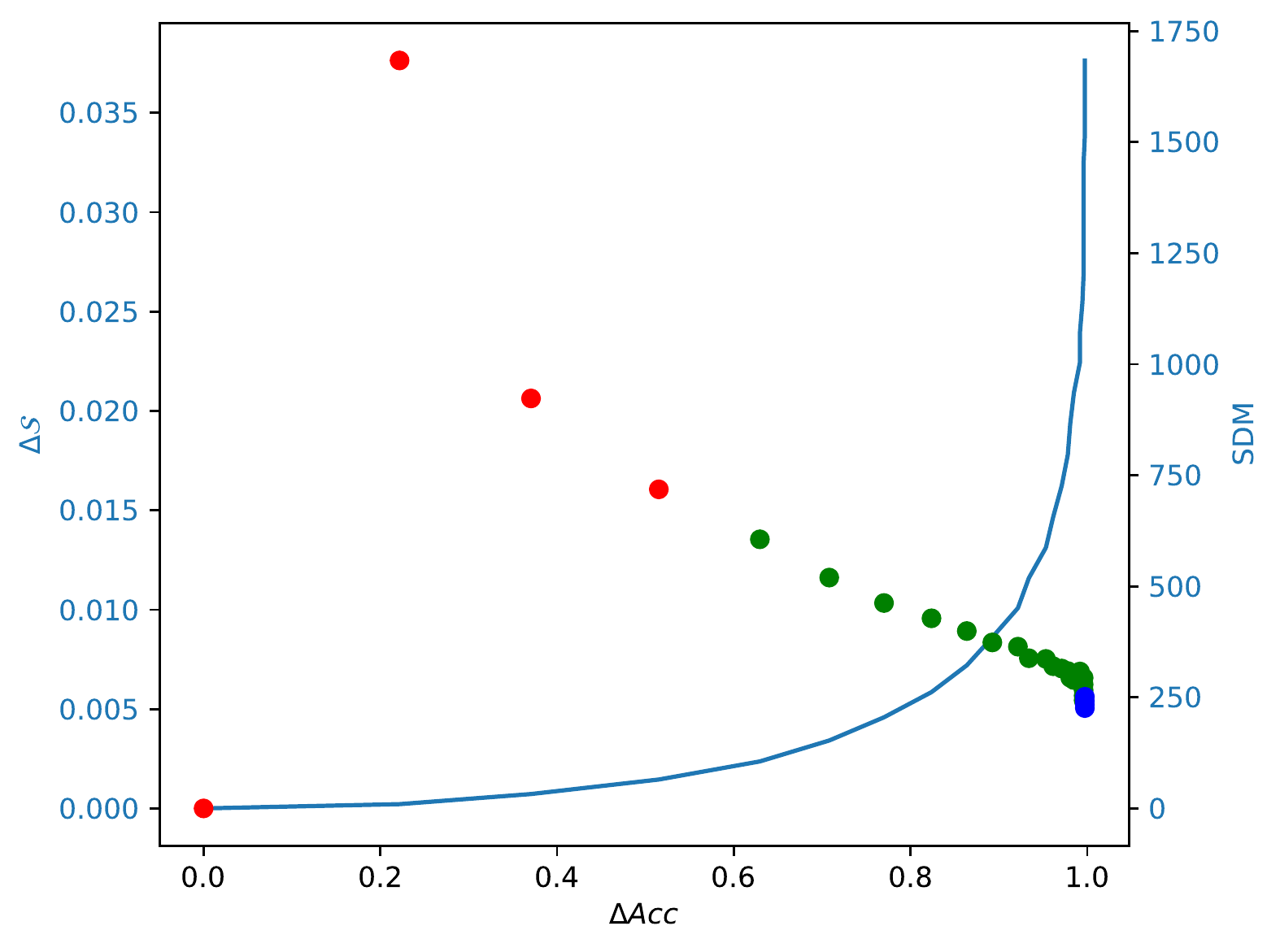}
    \caption{SDM score with unstable region marked in red and blue respectively.}
    \label{SDM_rb}
\end{figure}

\begin{figure}[htbp]
  \centering
  \subfigure{
  \begin{minipage}[t]{0.45\linewidth}
  \centering
  \includegraphics[width=\linewidth]{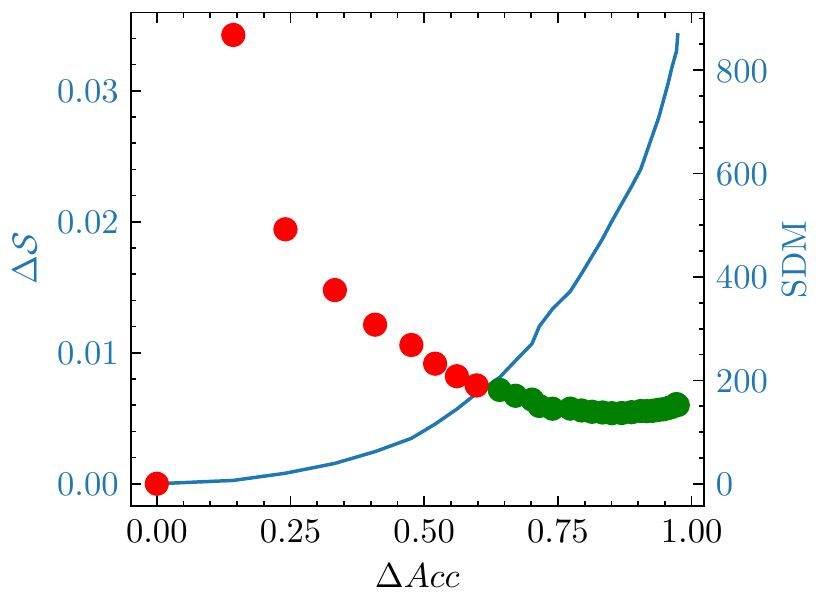}
  \end{minipage}%
  }%
  \subfigure{
  \begin{minipage}[t]{0.45\linewidth}
  \centering
  \includegraphics[width=\linewidth]{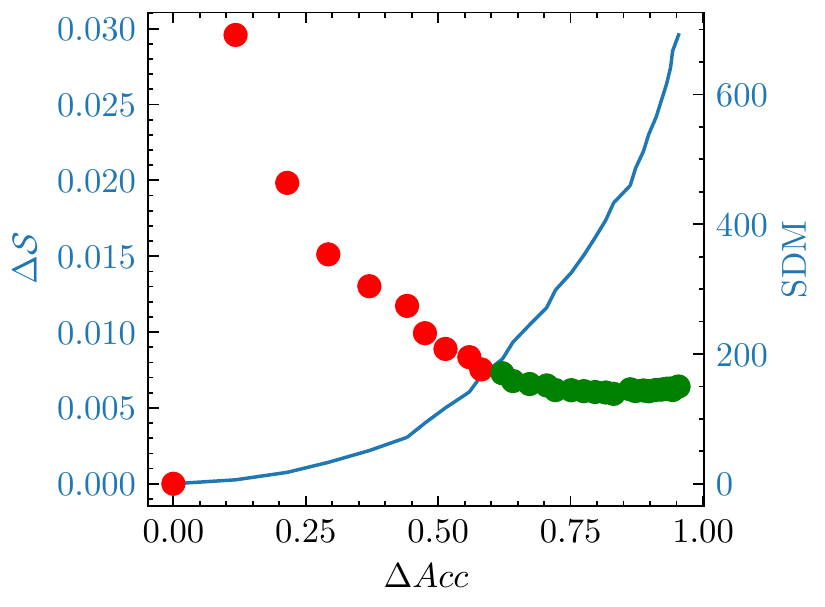}
  \end{minipage}%
  }%
  \centering
  \caption{stable SDM score without sudden decay under different target models.}
  \label{fig:side:a}
\end{figure}

As illustrated in figure \ref{SDM_rb}, we show the $SDM$ score of a perturbed image with different target models, but with same encoder-decoder structure and dataset. We can see that the $SDM$ score is highly stable in a region of $\Delta Acc\in (0.6,0.995)$(denoted as $\mathcal{L}$), which is marked in green. $SDM$ score are unstable with smaller or greater $\Delta Acc$, which is marked as $red$ and $blue$ respectively. However, we argue that most adversarial attempt (as shown in \ref{fig:side:a}) should falls in $\Delta Acc \in \mathcal{L}$. Too low or too high $\Delta Acc$ means the perturbation is either too weak or over-fits the datasets. Thus we argue that $SDM$ is a stable metric for adversarial attacks.

\section{Method}
\subsection{Framework}
\begin{figure}[htbp]
    \centering
    \includegraphics[width=\linewidth]{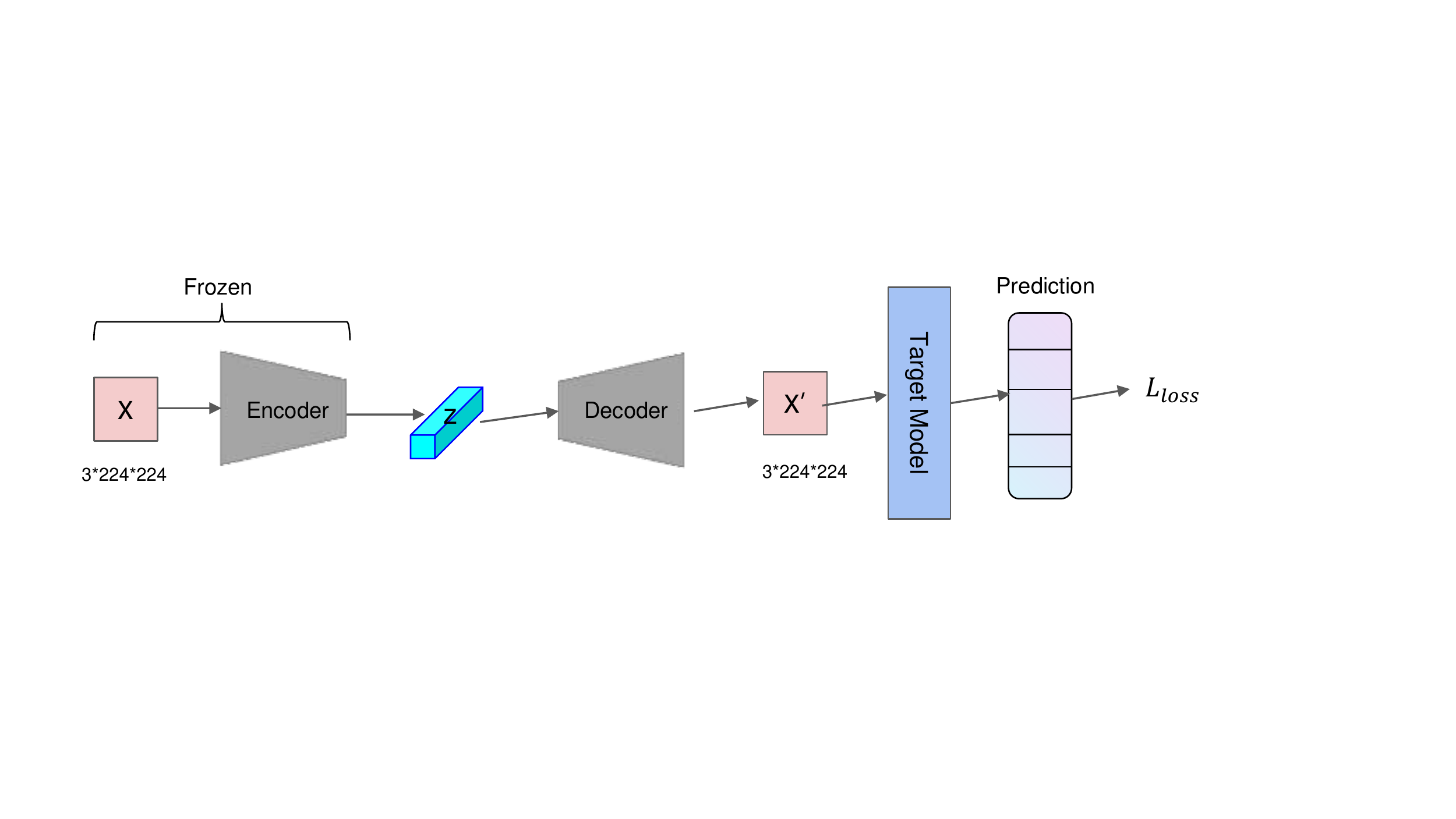}
    \caption{The framework of our method.}
    \label{framework}
\end{figure}

We adopt the pre-trained VAE from Stable Diffusion Model as our VAE module. For a dataset $X$, we encode it into $z=\mathcal{E}(X)\in \ZSEM$, then decode it back into pixel space as $X'=\mathcal{D}(z)$. We then feed $X'$ into the target model and calculate the loss function $\mathcal{L}$, backpropagate the loss to the latent space and update the latent variable $z$ by gradient ascent. We repeat this process for $T$ step with a learning rate $lr$. The algorithm for the noising phase is shown in Algorithm \ref{alg:noise}. The sketch of our framework is illustrated in figure \ref{framework}.

\begin{algorithm}[H]
  \caption{Noising Phase}
  \renewcommand{\algorithmicrequire}{\textbf{Input:}}
	\renewcommand{\algorithmicensure}{\textbf{Output:}}
  \label{alg:noise}
    \begin{algorithmic}
    \REQUIRE $X$, $\mathcal{F}$, $\mathcal{E}$, $\mathcal{D}$, $\mathcal{L}$, $T$, $lr$
    \ENSURE $X'$
    \STATE $z \leftarrow \mathcal{E}(X)$
    \FOR{$t=1$ to $T$}
    \STATE $X' \leftarrow \mathcal{D}(z)$
    \STATE $loss \leftarrow \mathcal{L}(\mathcal{F}(X'))$
    \STATE $z \leftarrow z + lr \cdot \nabla_z loss$
    \ENDFOR
    \STATE $z' \leftarrow z$
    \STATE $X' \leftarrow \mathcal{D}(z')$
  \end{algorithmic}
\end{algorithm}

\subsection{Loss Function and Evaluation Metrics}
We use the most common loss function for adversarial attack, the cross entropy loss(denoted as $\mathcal{L}_{ce}$). As for evaluation metrics, we use our $SDM$ metric to evaluate the quality of our adversarial attack. We also apply $SDM$ to traditional attacking methods by setting $\mathcal{S}_\mathcal{M}=1$ and compare them with our method.
\section{Experiments}
\subsection{Implementation Details}
We use the VAE from Stable Diffusion Model v2.0. As for dataset, we choose the validation set of ImageNet-1k. Due to our limited time and resources, in practice we randomly choose 1000 images from the validation set as our dataset. We choose several classification models as our target model including FocalNet, ConvNext, ResNet101 and ViT\_B. For each image, we first encode it into the latent space, then we do one forward, one backpropagation and do gradient ascent on the latent variable with a fixed $lr$. For each image we run 30 iterations of gradient ascent, as we find that the attack rate nearly converges after 30 epochs. For all PGD attack we use in the experiments, we set the number of iterations to 50, as we found that the loss of PGD attack converges after 50 iterations. We set $lr=5e-3$ for every attack methods.
\subsection{Results}\label{results} 

\subsubsection{Attacking The Target Models}\label{transfer}
As found in \cite{ViTAttack}, the family of ViT\cite{VIT} differs from the traditional CNN architectures when facing adversarial attacks. So we choose models from both family and test their performance.

We also apply traditional iteration/single step methods to attack the models and test their performance. We choose PGD\cite{PGD},FGSM\cite{goodfellow} and test their performance with our method, and proves that we can reach comparable performance with traditional methods. We set $\gamma = 0.999$ and $\varepsilon=1e-7 $ in our experiments. 

To quantify our adversarial examples, we use our $SDM$ metrics for evaluation. We report our top-1 accuracy and the corresponding $SDM$ score if our method and compare it to the traditional methods, as shown in table \ref{tab:transfer} and \ref{tab:sdm}.

Though our method has only achieve the best performance in terms of top-1 accuracy in the white-box setting when the target model is ViT\_B, we achieve the best $SDM$ score by about 10x compared to the traditional methods. This shows that our method is a stronger attack to the target model. However, it is not fair to compare the $SDM$ score of our method with the traditional methods, as the traditional method are not designed to maintain a high $\mathcal{S}$ like the semantic space does. But we can still see that the huge gap between the $SDM$ score of our method and the traditional method, which to some extent shows the superiority of our method.
\subsubsection{Cross-Model Transferability}
As the perturbation added on images is highly semantical, we expect the perturbation to be more transferable. We test our perturbation on different models but in the same dataset. The results are shown in table \ref{tab:transfer}.


  For the absolute attack rate, we can see PGD attack still outperforms others. However, our method achieve a very close performance to PGD attack, while being much transferable. Our method reaches the best performance in both the average attack rate and avarage $\Delta Acc$ except for the case when our training target model is ViT\_B. However, we have the best attack rate and the best $\Delta Acc$ in this case. The results indicate our method has the best performance in transferability comparing to the traditional method. As expected, perturbation crafted in semantic latent space is much more transferable than regular pixel space-based attacks. 

  \begin{table}[htbp]\label{tab:transfer}
    \centering 
    \tiny
    
    \caption{Transferability comparison on classification with different models. Here we report the top-1 accuracy and average $\Delta Acc$, average accuracy of four target models. The higher the $\Delta Acc$ the better. For the white-box situation where the surrogate model is the same as the target model, we set the background to be gray. The best results are bolded. All the traditional iteration/single step method are within the perturbation budget of $l_\infty<10$.  }
  \begin{tabular}{@{}ccccccl@{}}
  \toprule
                 & ConvNext\_base                          & ResNet101       & FocalNet\_base                          & ViT\_B                                  & Average $\Delta Acc$ & Average Acc     \\ \midrule
  Clean          & 0.8370                                  & 0.8340          & 0.8470                                  & 0.7410                                  & N/A                  & 0.8148          \\
  PGD(ConvNext)  & \cellcolor[HTML]{9B9B9B}\textbf{0.0110} & 0.5570          & 0.4070                                  & 0.6350                                  & 0.4946               & 0.5397          \\
  FGSM(ConvNext) & \cellcolor[HTML]{9B9B9B}0.3980          & 0.6030          & 0.5160                                  & 0.6140                                  & 0.3409               & 0.5325          \\
  \midrule
  Ours(ConvNext) & \cellcolor[HTML]{9B9B9B}0.0200          & \textbf{0.3900} & \textbf{0.2590}                         & \textbf{0.5480}                         & \textbf{0.6158}      & \textbf{0.3042} \\
  \midrule
  PGD(FocalNet)  & 0.4588                                  & 0.6150          & \cellcolor[HTML]{C0C0C0}\textbf{0.0190} & 0.6310                                  & 0.4619               & 0.4309          \\
  FGSM(FocalNet) & 0.5590                                  & 0.6380          & \cellcolor[HTML]{C0C0C0}0.4220          & 0.6260                                  & 0.3060               & 0.5613          \\
  \midrule
  Ours(FocalNet) & \textbf{0.2790}                         & \textbf{0.4190} & \cellcolor[HTML]{C0C0C0}0.0300          & \textbf{0.5310}                         & \textbf{0.6031}      & \textbf{0.3148} \\
  \midrule
  PGD(ViT\_B)    & 0.7630                                  & 0.7240          & 0.7660                                  & \cellcolor[HTML]{C0C0C0}0.0030          & 0.3279               & 0.5640          \\
  FGSM(ViT\_B)   & 0.6350                                  & 0.6110          & 0.6350                                  & \cellcolor[HTML]{C0C0C0}0.0630          & 0.3560               & \textbf{0.4860} \\
  \midrule
  Ours(ViT\_B)   & 0.6690                                  & 0.6670          & 0.6730                                  & \cellcolor[HTML]{C0C0C0}\textbf{0.0020} & \textbf{0.4009}      & 0.5026          \\ \bottomrule
  \end{tabular}
  \end{table}

  \begin{table}[htbp]
    \caption{We report the $SDM$ score with the highest attack rate of our method and  traditional methods on different models. The higher the $SDM$ score the better. The top 3 $SDM$ score is bolded.}
    \label{tab:sdm}
    \centering
    \tiny
    \begin{tabular}{|l|l|l|l|l|l|l|}
    \hline
              & PGD(ConvNext) & Ours(ConvNext) & PGD(FocalNet) & Ours(FocalNet) & PGD(ViT\_B) & Ours(ViT\_B) \\ \hline
    SDM score & 22.48         & \textbf{153.8}          & 14.78         & \textbf{149.8}          & 21.68      & \textbf{225.3}       \\ \hline
    \end{tabular}
    \end{table}
\section{Future Work}
In this paper we purpose a basic iteration-based framework to add perturbation on latent variables. However, we believe that more advanced update technique for latent variables can be explored. Future researches may works on more efficient and achieving ways to update latent variables, or try generator-based methods on latent space. We also show that the performance of attacks in latent space strongly depends on the quality of the latent space, which is determined by the encoder-decoder structure. Therefore, future researches may explore new encoder-decoder structure to improve the attack rate. As for our evaluation metrics $SDM$, it still faces the problem of sudden decay when attack rate is to high. Thus, future works may modify and improve the $SDM$ metric to enhance the stability.
\section{Conclusions}
In this paper, we explain the reasonability and neccessity to add perturbation in the semantic latent space. We also propose a basic iteration-based framework to add perturbation on latent variables on a highly semantical latent space created by a pre-trained VAE module of Stable Diffusion Model. As adversarial examples created in the latent space is hard to follow the $l_p$ norm constraint, we purpose a novel metric named $SDM$ to measure the quality of adversarial examples. To the best of our knowledge, \textbf{$SDM$ is the first metric that is specifically designed to evaluate adversarial attacks in latent space}. We also show that $SDM$ is highly model-independent and dataset-independent, while maintaining a high level of stability. We also test the cross-model transferability of perturbations crafted in latent space and proves their comparable performance with traditional methods. We believe that our $SDM$ metric can help future researches to evaluate the strength of latent spaces attack with a quantitive measure, thereby promoting the researches on latent space attack. Most importantly, we give an illustration of the potential of adversarial attacks in the semantic latent space, and \textbf{we believe that adversarial attacks in such highly semantical space can be a promising research direction in the future}.

\medskip

\bibliographystyle{plain}

\bibliography{ref}

\section*{Acknowledgements}
Thanks to all of you! I'll fix the appendix later. And the acknowledgements if available :D.



\end{document}